# Handwritten Digit Recognition by Elastic Matching

Sagnik Majumder[1*], C. von der Malsburg[2], Aashish Richhariya[3], Surekha Bhanot[4]

[1,3,4] Electrical and Electronics Engineering Department, Birla Institute of Technology and Science, Pilani, Rajasthan, India.
[2] Frankfurt Institute for Advanced Studies, Frankfurt, Germany.

* Corresponding author. Tel.: +91 8902339856; email: sagnikmjr2002@gmail.com

**Abstract:** A simple model of MNIST handwritten digit recognition is presented here. The model is an adaptation of a previous theory of face recognition. It realizes translation and rotation invariance in a principled way instead of being based on extensive learning from large masses of sample data. The presented recognition rates fall short of other publications, but due to its inspectability and conceptual and numerical simplicity, our system commends itself as a basis for further development.

## 1. Introduction

Deep learning (DL), see for instance [1], constitutes a breakthrough in pattern recognition, making it possible to recognize with very high reliability handwritten digits [2] or 1000 object types from natural photos [3]. Crucial for success are hierarchies of trained feature detectors that can test for the simultaneous presence of an array of sub-features (ultimately of pixels) while allowing for some flexibility in their spatial arrangement. A weakness of DL is the lack of full exploitation of transformation properties of objects in images. Enforcing shift symmetry in convolutional neural networks (CNN) only partially addresses this issue, as translation invariance has to be learned separately for each object. This weakness is made evident by the data augmentation methodology, the inflation of the input data set with large numbers of shifted or deformed versions of each input object, as described, for instance, in [1] or [2].

Here, a radically different approach [4] is being applied to handwritten digit recognition, implementing transformation invariance in a principled and object-independent way. The radical difference of our approach to Deep Learning is that the input figure is matched directly to models, using as hidden variables not excitation levels of neural feature neurons but links between image and model pixels, see Figure 1. In order to cope with substantial differences in the shapes of individual digits, multiple models for each digit are stored (and matched too). The quality of each trial match is evaluated in terms of a cost function containing the negative of the sum of point-to-point similarities and a measure of deformation of the map, see (4). The model matching the input digit with the lowest cost is taken as the winner. Since the process of match-finding allows for an arbitrary shift, rotation and small deformation, our system addresses invariance to these transformations in a principled way.

## 2. Related Work

In [5], the authors highlight the importance of parameter tuning in Gabor filters and its importance in the recognition performance of the filters. They report mathematical relationships between pairs of parameters and vary only the independent parameters for their experiments. The results that they report are comparable to the state-of-the-art results achieved by DL. In [6], the authors not only emphasize the importance of the correct choice of parameters for good classification performance but also stress the use of the correct classifier or distance measure for the final classification. To the best of our knowledge, these are the only recent publications which have dealt with MNIST digit classification through the direct use of Gabor filters and have achieved comparable classification results.

## 3. Gabor Features

Following a general trend (e.g., [7]), image and model pixels are labelled as jets $\Im_{v\theta}(x,y)$ of Gabor wavelets, using as convolution kernel

$$\psi_{v\theta}(\vec{x},\vec{f}) = \frac{1}{2\pi\sigma^2} e^{-\left[\frac{\vec{x}^2}{2\sigma^2}\right]} e^{i2\pi\vec{f}_{v\theta}\vec{x}} \qquad (1)$$

Here, $\vec{x}$ refers to image coordinates in units of pixel, $\vec{f}$ is the frequency (wave vector) of the oscillatory part, $v$ is the index of the frequency level, $\theta$ is the orientation of the wave vector and $\sigma$ is the width of the Gaussian envelope of the convolution kernel. Two frequency levels, $0.25/pixel$ and $0.125/pixel$, are used which are calculated with $\mu$, the ratio of neighboring frequencies, and $f_o = 0.5/pixel$ using $f_v = f_o \mu^{-v}$ such that $v = \{1,2\}$, four equally spaced orientations in the range $(0,\pi)$ and $\sigma$ of value $2\pi$. Let $I(\vec{x})$ be the grey-level of an image at position $\vec{x}$ and feature values are defined by the convolution

$$\Im_{v\theta}(x,y) = |\sum_{x'} \psi_{v\theta}(\vec{x}-\vec{x}')I(\vec{x}')|. \qquad (2)$$

The vertical bars signify that the absolute value of the complex wavelet components is taken. The sum runs over all pixels of the image. These feature vectors, each of which has $8 = 2\times 4$ components, are used as labels of the image pixels. They are normalized to length 1 and the results are referred to as "jets":

$$\vec{\Im}(\vec{x}) = \frac{\{\Im_{v\theta}(\vec{x})\}}{\|\{\Im_{v\theta}(\vec{x})\}\|} \ . \qquad (3)$$

## 4. Pattern Matching

In order to evaluate the similarity between the patterns in a test image $T$ and a model image $M$, feature arrays in both images are interpreted as graphs $G = (V,E)$ with vertices or "nodes" $i \in V$ corresponding to the image sampling points labelled with jets, and links $(i,j) \in E$ labelled with difference vectors $\Delta_{i,j} = \vec{x}_i - \vec{x}_j$ of distances between nodes $i$ and $j$ within the image. Given that Gabor magnitudes, (2), vary slowly as function of pixel position, our image graphs $G$ were limited to a grid of $10\times 10$ nodes spaced by 2 pixels in each dimension, and all links between vertical or horizontal neighbors (that is, a node in the interior of the graph has four links, border nodes having fewer links) are included in the vertex set $E$. The two graphs $G^T$ and $G^M$ are compared in terms of the mapping between corresponding points, see Fig. 1. During the matching procedure for a given test image the model graph $G^M$ is kept fixed and the test graph $G^T$ is varied by moving it rigidly pixel by pixel over the test image (test graph nodes picking up positions $\vec{x}_i$ and jets $\Im_i^T$ where they come to fall) such that the test graph checks all possible positions to account for the shift. During this movement of the graph, the test image is considered to have periodic boundary conditions, such that the graph is wrapped around it. In each position, the

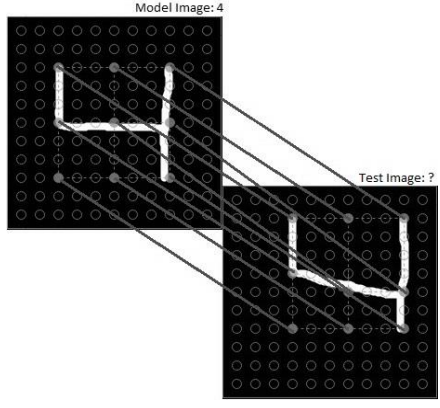

Fig. 1. Digit recognition by elastic matching of test images (T) to model images (M) of known digit identity (schematic). See last paragraph of section 4. for detailed explanation.

trial graph is evaluated in terms of the cost function

$$C(G^T) = \lambda C_e + C_v$$
$$= \lambda \sum_{i,j \in E} S_e(\Delta_{i,j}^T, \Delta_{i,j}^M) - \sum_{i \in V} S_v(\mathfrak{I}_i^T, \mathfrak{I}_i^M) \quad (4)$$

Here, $\lambda$ is a parameter for the relative weight of the two similarity terms:

$$S_e(\Delta_{i,j}^T, \Delta_{i,j}^M) = (\Delta_{i,j}^T - \Delta_{i,j}^M)^2 \quad (5)$$

is measuring the distortion between the corresponding links and

$$S_v(\mathfrak{I}_i^T, \mathfrak{I}_i^M) = \mathfrak{I}_i^T \cdot \mathfrak{I}_i^M, \quad (6)$$

the scalar product of the normalized jet vectors, is measuring feature similarity.

The match of a model to the test image proceeds by generating a test graph $G^T$ in identical structure and position to the model graph (or in a rotated version, see below) and then altering it in two phases called global move and local move. During the global move the graph is moved rigidly over the test image in steps of size 1 pixel to find the position with minimal cost. (During that phase the deformation term $C_e$ of the cost function is zero, of course.) In the local move, individual node points in the test graph are allowed to move over a restricted range (one grid point right and left, up and down) in search of a better cost value, each grid point visited once. In series of experiments, the relative weight parameter was varied and the optimal value of $\lambda = 3 \times 10^{-9}$ was settled for.

Many of the MNIST digits are slanting to the side, a regular feature of hand-written characters. This is being dealt with by estimating this slant, as described in [8], as the second moment

$$m = \frac{<xy> - <x><y>}{<y^2> - <y>^2} \quad (7)$$

where $<z>$ is the weighted average of the quantity $z$ ( $z$ standing for $x$, $y$, $xy$ etc.), using as weights the pixel grey levels normalized across the image to the sum 1. This value $m$ is taken as the tangent of the orientation of the digit, which is a good approximation if the orientation is not too far from the vertical. This orientation is used to rotate the wave vector $f$ in the Gabor convolution kernel, (1) to extract jets at each pixel according to (2) and to rotate the standard $10 \times 10$ sample grid over the model resp. over the test image, using wrap-around boundary conditions.

Fig. 1. portrays a sample matching procedure. Pixels (circles) are labelled with low-level feature jets (Gabor jets). The model digit is represented by a model graph (filled circles and dashed lines) and is compared to the graph that is first moved with identical shape over the test image ("global move") and after finding the optimal position, is deformed ("local move"). Graphs are compared in terms of a cost function evaluating the similarity of features

picked up at corresponding nodes (connected by oblique lines, the dynamic links) and evaluating the deformation. All model digits are matched to the test digit and the one with the lowest cost is selected.

## 5. MNIST Dataset

MNIST [10] dataset is a dataset of handwritten Arabic numerals and is popularly considered as a standard dataset for handwritten digit classification in optical character recognition and machine learning research. MNIST stands for modified National Institute of Standards and Technology. The MNIST dataset is made out of the original NIST dataset; thus, modified NIST or MNIST. There are 60,000 training images (some of these training images can be used for cross-validation purposes) and 10,000 test images, both drawn from the same distribution. All these gray scale images are size normalized, and centered in a fixed size image where the center of gravity of the intensity lies at the center of the image with $28 \times 28$ pixels. The part of the image that contains the actual Arabic numeral is the central $20 \times 20$ part.

## 6. Experiments and Results

Our system is built solely in Python, using the package 'bob.ip.gabor' [9]. To test our system, a model set of 40,000 images and a test set of 8,000 images from the MNIST dataset are used. The rationale for not using all 60,000 MNIST training images in the model set and all 10,000 testing in the test set is the inequality in the number of images per digit in both training and testing sets of MNIST. As the digit with the minimum number of training images has a few more than 4,000 entries, an equal number of 4,000 model images for each digit from the MNIST training set and 800 testing images for each digit from the test set are randomly selected.

Experiments on the model and the test sets, as mentioned above, are carried out using all combinations of parameters from the sets: $N_x \in \{5,10,20\}$, $N_d \in \{4,8\}$, $\mu \in \{\sqrt{2}, 2, 2\sqrt{2}\}$ and $\lambda \in \{3 \times 10^{-9}, 3 \times 10^{-6}, 3 \times 10^{-3}\}$. Here, $N_x$ is the number of image sampling points per dimension of the graphs, both model and test, $N_d$ is the number of equally spaced wave vector orientations in the range $(0, \pi)$ and $\mu$ is the ratio of neighboring Gabor frequency levels. The optimum set of parameters is found as $\{N_x = 10, \mu = 2, N_d = 4, \lambda = 3 \times 10^{-9}\}$. This set of parameters gives a recognition accuracy of $96.528\%$. When the system is run with other parameter sets, it is very sensitive to the parameter $\lambda$, bringing the recognition rate down to $41.163\%$ for $\lambda = 3 \times 10^{-3}$, at which the sample grid is very stiff, illustrating the importance of match deformation. The other parameters affect the recognition accuracy much less, recognition rates remaining in the range $[90.217\%, 96.528\%]$. Table 1. shows the effect of different parameter combinations on the classification results. Results for $\mu = 2\sqrt{2}$ and $\mu = \sqrt{2}$, and $\lambda = 3 \times 10^{-3}$ are excluded because the recognition accuracies are comparatively bad, in the range $[90.217\%, 91.637\%]$.

To demonstrate the translation invariance of our system a separate experiment (with the optimal parameters given above) is done in which the test images were modified by randomly shifting the digits in the test images by up to half of the image diameter (using periodic boundary conditions such that the digits may wrap around the edges and the corners), see for instance Fig.1, where the digit '4' is shifted from its original position at the center of the image. The original padded MNIST images are $28 \times 28$, so to implement the shift the central $20 \times 20$ part containing the actual digit is taken and shifted. The new position of the digit in the test image is found by the test graph during the global move. The model images are kept the same as earlier. This gives a recognition accuracy of $96.364\%$ which shows that our system is fully translation invariant, in addition to the deformation and rotation invariance shown in the main experiment.

To counter the criticism that our system needs excessive numbers of model digits, the above set of 40,000 model images is reduced to 1,000 model images, by randomly picking 100 images for each digit from the above set, continuing to work with the original 8,000 test images (leaving them non-shifted). With these model and test sets,

Table 1. Effect of Parameter Combinations on Recognition Results

| $N_x$ | $\mu$ | $N_d$ | $\lambda$ | Recognition Accuracy |
|---|---|---|---|---|
| 5 | 2 | 4 | $3\times10^{-9}$ | 93.826% |
| 5 | 2 | 4 | $3\times10^{-6}$ | 93.008% |
| 5 | 2 | 8 | $3\times10^{-9}$ | 93.779% |
| 5 | 2 | 8 | $3\times10^{-6}$ | 93.565% |
| 10 | 2 | 4 | $3\times10^{-9}$ | 96.528% |
| 10 | 2 | 4 | $3\times10^{-6}$ | 94.892% |
| 10 | 2 | 8 | $3\times10^{-9}$ | 95.996% |
| 10 | 2 | 8 | $3\times10^{-6}$ | 95.312% |
| 20 | 2 | 4 | $3\times10^{-9}$ | 95.110% |
| 20 | 2 | 4 | $3\times10^{-6}$ | 93.965% |
| 20 | 2 | 8 | $3\times10^{-9}$ | 94.125% |
| 20 | 2 | 8 | $3\times10^{-6}$ | 94.008% |

a recognition accuracy of $96.109\%$ is achieved.

## 7. Discussion

In comparison to other systems, especially DL systems, the present model stands out by numerical and conceptual simplicity, inspectability and great potential for further development. However, some weaknesses in its present form must be admitted. Our recognition rates fall short of published DL results, e.g., $\sim 99.8\%$ [11], and our system is based on an iterative procedure of matching to multiple models per digit, so that individual recognition events take more time than in feed-forward based DL systems. It can be argued, however, that the system was built with minimal effort as a summer project (of SM), and its simplicity is seen as essential for the demonstration of a set of concepts that have the potential to go significantly beyond DL, as it will be argued presently.

The main deviation of our approach from current neural systems, including DL, is its emphasis on dynamic links. The core of the system is the match process between input and model. It is based on lowest-level features (Gabors in our present version). The decisive aspect of these features is not high pattern-specificity (as in the topmost CNN layer in DL systems) but their spatial arrangement (represented by lateral links within the image or model representation). In the present version, this arrangement is formulated in two stages, the grouping of concentric Gabors into jets and the spatial arrangement of jets in input and model. The grouping of Gabors into jets is the basis for the computation of similarities between tentative correspondence points in input and model, (6). The spatial arrangement of jets in input and model is important for constraining the shape of the mapping between the two images being compared, see the distortion term, (5).

The dynamic links constituting the mapping between input and model bases the pattern recognition on homeomorphism, "the same features in the same arrangement in input and model". In this statement, there is no mention of the position, size or orientation of the input pattern, as it shouldn't. Accordingly, our system implements invariant recognition in its proper natural way.

The decisive advantage of neural systems is their amenability to learning and self-organization. This advantage is not lost with the introduction of dynamic links, as dynamic links can be implemented in a fully neural system, see [12], and dynamic constraints on the arrangement of links in admissible input-model mappings can be learned from examples [13], [12].

As mentioned, a weakness of our system is that it uses many models per digit to be recognized. This is particularly awkward as it necessitates, in the present form of the system, a time-consuming sequence of individual input-model matches, each realized iteratively. Several measures are available to attack this problem. First, our (randomly selected) set of model digits can be reduced in size by weeding out those that don't contribute to recognition in sets of trial runs. Second, the system could be based on a single dynamic link map that develops from coarse to fine during the recognition process along with a coarse-to-fine search through the model database. The feasibility of this has been shown in [14], and our treatment of digit slant runs exactly along those lines. Third, although it is unavoidable to store qualitatively different variants of a digit (such as '1' with or without upstroke, '7' with or without a dash through the middle), many variants are just elastic deformations of each other. Such deformation is being coped with using our local move, but a system learning and applying digit-specific deformation patterns could work with one model per qualitative variant of each digit.

It is hoped that the essential message implicit in our little model is not obscured by its obvious temporary weaknesses. This message is that by putting emphasis on links (instead exclusively on feature decision units) it becomes possible to separate out the representation of *pattern structure* on the one hand from the laws of *pattern transformation* on the other, patterns being represented as linkages of generic low-level feature units, transformations as regular sets of dynamic links. This separation makes it possible for the system to acquire novel models from single examples and to apply general transformation laws to them immediately.

## 8. Conclusion

The present model for handwritten digit recognition is a computer-adaptation of a neural theory of invariant recognition. This theory fundamentally deviates from current neural theories by having link variables in addition to neural signals. This opens the possibility of basing invariant recognition on the homeomorphism between the input figure and explicit models, homeomorphism meaning "same elements in the same arrangement". The approach obviates the learning from massive data and can work with moderate numbers of models.

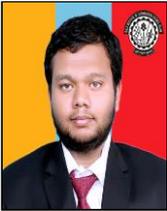
**Sagnik Majumder** was born in 1996 in Kolkata, India. He is an undergraduate student at Birla Institute of Technology and Science Pilani, India with a major in electronics and instrumentation engineering.

He was a Research Intern under the supervision of Prof. C. von der Malsburg in the Department of Neuroscience at Frankfurt Institute for Advanced Studies, Germany during the summer of 2017. He has previously published a conference paper on temperature compensation of ISFET-based pH meter using Neural Networks in IEEE Regional Symposium of Microelectronics 2017, Malaysia. He is mainly interested in research in the field of Computer Vision, Machine Learning and Deep Learning and is associated with the Multimedia Research Laboratory of his University.

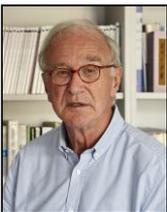
**Christoph von der Malsburg** received his diploma in physics from the University of Heidelberg, Germany in 1968 and his PhD in physics from the same institution in 1970.

He acted as Research Scientist in neuroscience at a Max-Planck Institute in Göttingen, Germany, as Professor of Computer Science at the University of Southern California in Los Angeles, USA, and Professor of Systems Biophysics at Ruhr-University Bochum, Germany. His present position is Senior Fellow at the Frankfurt Institute for Advanced Studies, Frankfurt, Germany. His research interest in the past has been the ontogenesis of ordered connectivity patterns in the vertebrate visual system, and he presently works on implementing a coherent functional theory visual perception.

Prof. von der Malsburg is Fellow of the International Neuroscience Society, from which he received the Hebb Award 2003. He had also received the 1994 Pioneer Award of the Neural Network Council of the IEEE. He has co-founded two companies on the basis of his theory of vision.

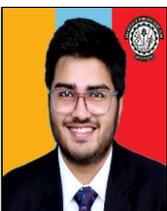
A**ashish Richhariya** was born in 1994 in India. He is an undergraduate student at Birla Institute of Technology and Science Pilani, India with a major in electronics and instrumentation engineering.

He is expected to graduate in May 2018 and get a degree in Bachelor of Engineering (Honours). He is mainly interested in research in the field of Computer Vision, Artificial Intelligence, and Robotics. He has been working with the University's research group for Robotics and has been involved in a number of important projects for that group.

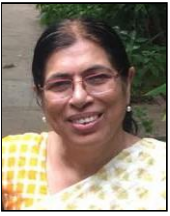 **Surekha Bhanot** received her bachelors in mechanical engineering in 1979, masters of philosophy in instrumentation in 1983 from Birla Institute of Technology and Science(BITS) Pilani, India and her PhD from IIT Roorkee, India in 1995.

She has assumed various academic positions in BITS Pilani and TIET, Punjab, India since 1979. Her present position is Professor in the Department of Electrical and Electronics Engineering at BITS Pilani. Her research interest in the past has been the **Artificial Intelligence Applications in Soft Sensing, Process Modelling and Control**.

Prof. Bhanot is a life member at the Instrumentation Society of India, the Indian Science Congress, the Indian Society for Technical Education and the Museum of Science and Industry.